\documentclass[10pt,twocolumn,letterpaper]{article}

\usepackage{wacv}
\usepackage{times}
\usepackage{epsfig}
\usepackage{graphicx}
\usepackage{amsmath}
\usepackage{amssymb}
\usepackage{multirow}
\usepackage{makecell}
\usepackage[table]{xcolor}
\usepackage{fancyhdr}
\usepackage{arydshln}
\usepackage{animate}
\usepackage{array}
\usepackage{xr}

\usepackage[ruled,vlined]{algorithm2e}
\usepackage{times}
\usepackage{epsfig}
\usepackage{graphicx}
\usepackage{amsmath}
\usepackage{amssymb}
\usepackage[utf8]{inputenc} 
\usepackage[T1]{fontenc}    
\usepackage{url}            
\usepackage{booktabs}       
\usepackage{amsfonts}       
\usepackage{nicefrac}       
\usepackage{microtype}      
\usepackage{xcolor}         
\usepackage{microtype}      
\usepackage{graphicx}
\usepackage{amsmath}
\usepackage{mathtools}
\usepackage{makecell}
\usepackage{amsthm}
\usepackage{enumitem}
\usepackage{subcaption}
\usepackage{booktabs}
\usepackage{ragged2e}
\usepackage{arydshln}
\usepackage[hang,flushmargin]{footmisc}
\usepackage{multirow}
\usepackage{multicol}
\usepackage{xr-hyper}
\usepackage[pagebackref=true,breaklinks=true,colorlinks,bookmarks=false]{hyperref}

\newtheorem{remark}{\textbf{Remark}}

\newcommand{\darkgray}{\rowcolor[gray]{.75}}
\newcommand{\verydarkgray}{\rowcolor[gray]{.60}}

\def\wacvPaperID{878} 

\wacvfinalcopy 

\ifwacvfinal
\def\assignedStartPage{9876} 
\fi

\ifwacvfinal
\usepackage[breaklinks=true,bookmarks=false]{hyperref}
\else
\usepackage[pagebackref=true,breaklinks=true,colorlinks,bookmarks=false]{hyperref}
\fi

\ifwacvfinal
\else
\fi

\begin{document}
\setlength{\abovedisplayskip}{3pt}
\setlength{\belowdisplayskip}{3pt}
\title{AuxAdapt: Stable and Efficient Test-Time Adaptation for Temporally Consistent Video Semantic Segmentation}
\author{Yizhe Zhang \thanks{Work done at Qualcomm AI Research.} \thanks{Nanjing University of Science and Technology, Nanjing, China} \\\small{yizhe.zhang.cs@gmail.com}
\and
Shubhankar Borse \thanks{Qualcomm AI Research, an initiative of Qualcomm Technologies, Inc.} \thanks{Equal contribution.}\\\small{sborse@qti.qualcomm.com}
\and
Hong Cai \footnotemark[3] \footnotemark[4]\\\small{hongcai@qti.qualcomm.com}
\and
Fatih Porikli \footnotemark[3]\\\small{fporikli@qti.qualcomm.com}}

\maketitle

\begin{abstract}
\vspace{-0.13in}
In video segmentation, generating temporally consistent results across frames is as important as achieving frame-wise accuracy. Existing methods rely either on optical flow regularization or fine-tuning with test data to attain temporal consistency. However, optical flow is not always avail-able and reliable. Besides, it is expensive to compute. Fine-tuning the original model in test time is cost sensitive. 

This paper presents an efficient, intuitive, and unsupervised online adaptation method, AuxAdapt, for improving the temporal consistency of most neural network models. It does not require optical flow and only takes one pass of the video. Since inconsistency mainly arises from the model’s uncertainty in its output, we propose an adaptation scheme where the model learns from its own segmentation decisions as it streams a video, which allows producing more confident and temporally consistent labeling for similarly-looking pixels across frames. For stability and efficiency, we leverage a small auxiliary segmentation network (AuxNet) to assist with this adaptation. More specifically, AuxNet readjusts the decision of the original segmentation network (Main-Net) by adding its own estimations to that of MainNet. At every frame, only AuxNet is updated via back-propagation while keeping MainNet fixed. We extensively evaluate our test-time adaptation approach on standard video benchmarks, including Cityscapes, CamVid, and KITTI. The results demonstrate that our approach provides label-wise accurate, temporally consistent, and computationally efficient adaptation (5+ folds overhead reduction comparing to state-of-the-art test-time adaptation methods).
\end{abstract}
\vspace{-0.21in}

\pagenumbering{gobble}
\section{Introduction}\label{sec:intro}
\begin{figure}[t!]
\vspace{-0.05in}
\centering
\includegraphics[width=0.99\linewidth]{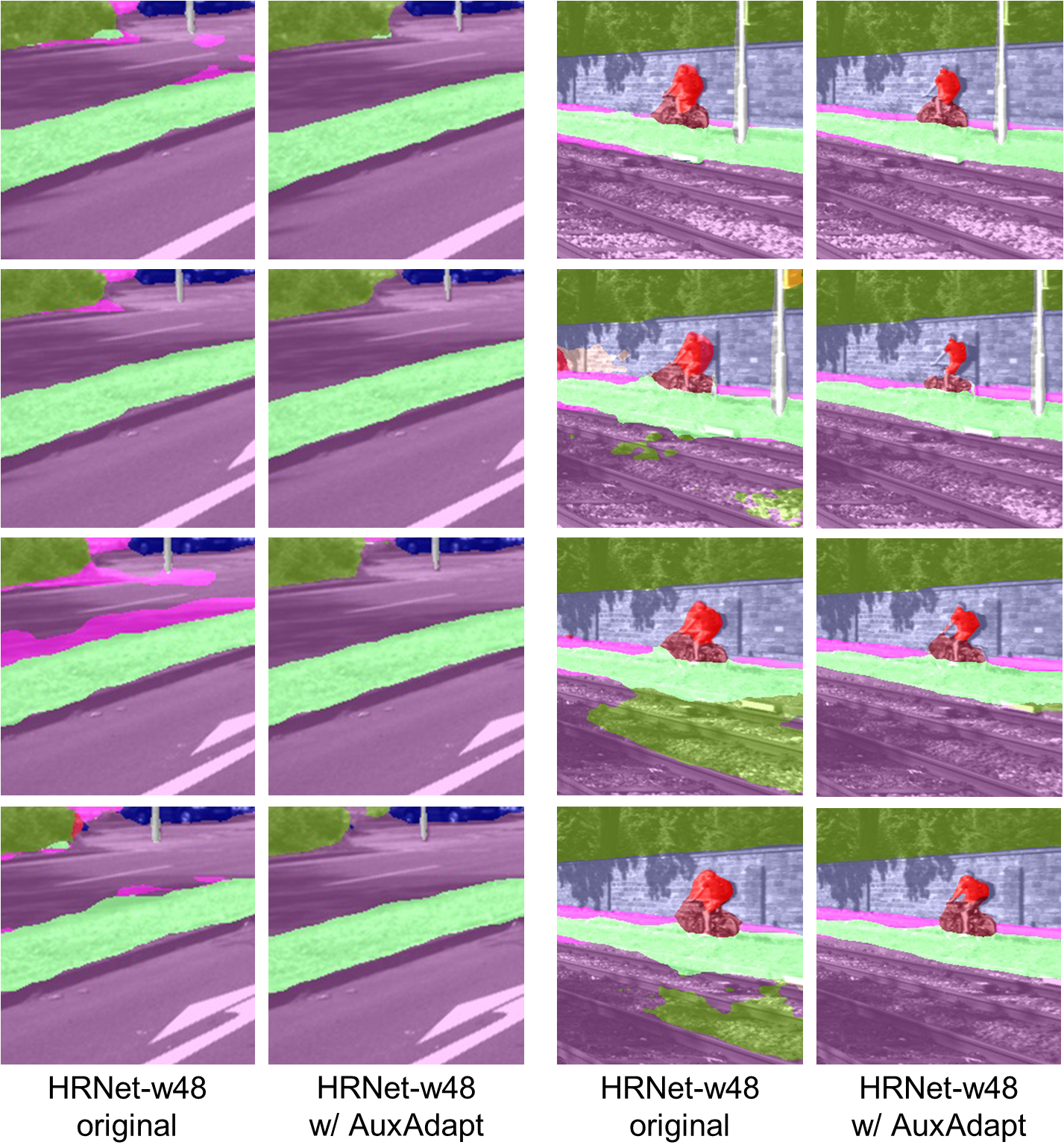}
\caption{Segmentations of two pairs of consecutive video frames. Columns 1 and 3 show the results of the state-of-the-art HRNet-w48 model~\cite{wang2020deep}. Columns 2 and 4 are obtained by applying our efficient AuxAdapt to HRNet-48. As visible, AuxAdapt improves the temporal consistency significantly. 
} 
\label{fig:illustration_tc}
\vspace{-0.2in}
\end{figure}

\vspace{-0.06in}
Recent years have witnessed remarkable progress in image-based semantic segmentation. With the rising popularity and pervasiveness of videos, there is now an increasing focus on video segmentation as a necessary functionality for higher-level computer vision tasks. While it is possible to treat video segmentation as an image segmentation problem and apply image-based models to each frame independently, such segmentations usually lack consistency in time. In other words, image pixels across consecutive video frames that belong to the same semantic class and share similar visual appearances can be labeled differently, resulting in artifacts such as flickering of segmentation. See examples in the 1st and 3rd columns of Fig.~\ref{fig:illustration_tc}.


There have been several attempts to bring temporal consistency in video semantic segmentation. Most methods utilize optical flow to impose consistency as it establishes pixel correspondence across frames. For instance, \cite{cheng2017segflow, ding2020every} propose a joint training of segmentation and optical flow, showing that the network can provide more consistent segmentation by learning both tasks together. Other methods incorporate pretrained optical flow as an additional channel of information to the network~\cite{nilsson2018semantic} for regularization during training~\cite{liu2020efficient}. While optical flow is commonly employed, there are drawbacks. First, inferring accurate optical flow maps is challenging; thus, such a dependency limits the performance of segmentation, often causing additional issues. Moreover, existing joint optical flow and segmentation methods need accurately annotated datasets. Besides, except for~\cite{liu2020efficient}, all these methods require the optical flow during inference, which creates computational overheads. 

Some previous approaches incorporate test-time model adaptation to enhance temporal consistency~\cite{bonneel2015blind, lai2018learning, miksik2013efficient, yao2017occlusion}. Nevertheless, most of these methods require dense correspondence through motion estimation or patch matching~\cite{barnes2009patchmatch}. A few works that do not use explicit optical flow computation, e.g.,~\cite{fayyaz2016stfcn, hu2020temporally}, still depend on feature conformity across multiple frames. In their pioneering work, Lei et al.~\cite{lei2020blind} have proposed Deep Video Prior (DVP), which can do away with the optical flow computation by leveraging how a deep neural network learns. Given a test video and the processed frames, DVP trains a network from scratch to generate these processed frames and employs early stopping to prevent the network from overfitting to temporally-inconsistent patterns. While DVP can provide adaptation, it demands training the original network model for more than $20$ epochs on each test video, resulting in a drastic computational cost.

\begin{figure*}[t!]
    \centering
    \includegraphics[width=1\linewidth]{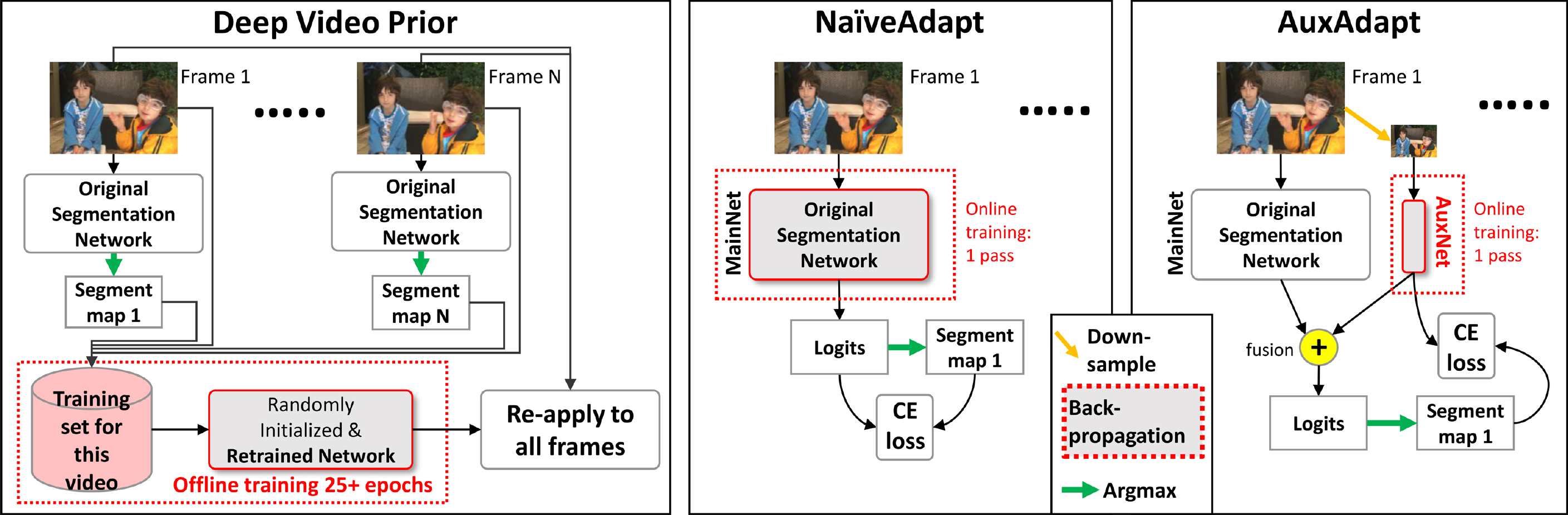}
    \caption{\textbf{Left:} Deep Video Prior (DVP)~\cite{lei2020blind} first applies the original network to all video frames. The outputs, together with the corresponding inputs, are collected to form a training set. A new network is then trained based on this set using at least 25 epochs. Finally, the retrained network is applied to the same video to obtain the final outputs. This means DVP applies twice inference over the entire set of frames and a full-scale, computationally expensive retraining.  \textbf{Middle:} NaiveAdapt: the main segmentation network is being updated when processing through the video frames. \textbf{Right:} AuxAdapt uses a tiny auxiliary network on lower spatial resolution together with the main segmentation network. It only updates AuxNet. This has several advantages (see Introduction). Note that AuxNet estimates are fused with MainNet \textit{before} cross-entropy loss with softmax (CE)~\cite{CEL}. Logits is a $H\times W \times K$ tensor.  
    } 
\label{fig:overview}
\vspace{-0.1in}
\end{figure*}

Our paper proposes an efficient unsupervised online (test-time) adaptation method to promote temporal consistency of video semantic segmentation. Our approach applies to any existing segmentation models. On a high level, we enable the network to efficiently learn from its own decisions without a dependency on optical flow or other temporal features either in training or test. The network adapts based on the (hard) labeling decisions it makes for the pixels of the current frame while it sequentially segments video frames. Our observation is that consecutive frames share similar visual content; therefore, by reinforcing the network to bolster on its previous responses, it is possible to obtain consistent responses for similar regions in future frames. Our intuition here is that we do not need to update the entire network to accomplish such a reinforcement. Instead, we incorporate a tiny auxiliary network (AuxNet) to steer and assist with the adaptation, as illustrated in Fig.~\ref{fig:overview} (right). We refer to our method as \textbf{AuxAdapt}.

%
    
    


Since this AuxNet can be easily trained with only one backward pass in back-propagation on a single frame, our method is very efficient. In comparison, existing adaptation methods such as DVP require training on a large set of frames for significantly more epochs. AuxNet retains a small architecture and works in lower spatial resolution and lower frame rate. It is trained for the same semantic segmentation task as the main segmentation network (MainNet), to which it provides test-time adaptation. During inference for each frame, MainNet is frozen, and only AuxNet is updated as the integrated model streams through the video. The final segmentation is determined by the ``aggregated" outputs of MainNet and AuxNet. In this way, the segmentation model adapts continuously to a given video with only a fraction of the computational cost of DVP since only AuxNet gets updated instead of the entire MainNet.

Furthermore, our proposed AuxAdapt maintains the segmentation accuracy while improving temporal consistency online. As MainNet is kept intact and contributes to the aggregated output, we prevent catastrophic forgetting and limit the variation of accuracy. Although AuxNet employs a smaller architecture and usually does not provide as high of a standalone accuracy as MainNet, training on the aggregated outputs allows it to distill knowledge from MainNet during the adaptation. As we shall see in the experiments, even when AuxNet's accuracy is a few percentage points lower than that of MainNet (in terms of mIoU), the overall segmentation accuracy is maintained. In some cases, the overall accuracy even improves slightly due to the ensemble effect of the two networks. In contrast, the state-of-the-art DVP trains a new network for a specific test video, and their model does not provide accurate segmentation for other test data without expensive retraining. Moreover, DVP relies on early stopping to a trade-off between temporal consistency and accuracy on the test video, but does not provide a clear criterion for early stopping in test time.

We summarize our main contributions as follows:
\vspace{-2mm}
\begin{itemize}
    \item We present an unsupervised online adaptation method, AuxAdapt, to boost temporal consistency of video semantic segmentation. Our approach does not require optical flow (or cross-frame features). It can be appended to any segmentation network.
\vspace{-2mm}
    \item We introduce a novel tiny auxiliary network (AuxNet) that accompanies the selected main segmentation network (MainNet). The final segmentation is obtained by fusing these two networks' outputs. Only AuxNet is updated at each frame with only one epoch during test time, which significantly reduces the computational load. AuxNet can take advantage of temporal and spatial sub-sampling for further acceleration.
\vspace{-2mm}
    \item We leverage on a simple change-detection-based adaptive momentum when performing the online adaptation, which adjusts the momentum coefficient based on the difference of two consecutive frames. We show that this provides a good balance between temporal consistency and segmentation accuracy.
\vspace{-2mm}
    \item We conduct extensive experiments on Cityscapes, Camvid, and KITTI with comprehensive ablation studies that demonstrate the efficacy of our approach. Overall, AuxAdapt consistently improves the temporal consistency of state-of-the-art models by several percentage points while maintaining segmentation accuracy. Notably, this is achieved with a tiny fraction of the computation cost required by existing methods (e.g., $<$2\% of what DVP incurs).
\end{itemize}

\vspace{-0.02in}
\section{Related Work}\label{sec:related work}
\vspace{-0.05in}

\noindent \textbf{Video Semantic Segmentation:}
Since the introduction of fully convolutional networks (FCNs) \cite{long2015fully}, various network architectures have been developed to improve the segmentation accuracy as well as efficiency, e.g., U-Net \cite{ronneberger2015u}, PSPNet \cite{zhao2017pyramid}, DeepLab \cite{chen2017deeplab}, and HRNet \cite{wang2020deep}, to count a few. Applying an image-based semantic segmentation network for videos in a per-frame fashion is a common choice in practice due to its simplicity. Although recent state-of-the-art segmentation networks (e.g., HRNet) can achieve impressive results on individual images, they often induce temporally inconsistent results when applied to video frames \cite{liu2020efficient}.

\vspace{0.01in}
\noindent \textbf{Temporal Consistency:} There are numerous studies on improving the temporal consistency of video processing tasks such as colorization, dehazing, and segmentation~\cite{lai2018learning, lei2020blind, miksik2013efficient}. More related to our paper are those that study temporal consistency of video semantic segmentation. In the existing literature, researchers have proposed incorporating different information during training to improve segmentation temporal consistency, e.g., optical flow~\cite{hur2016joint, nilsson2018semantic}, 3D structure~\cite{floros2012joint, kundu2014joint}, and/or utilizing more complex models, e.g., Recurrent Neural Network (RNN)~\cite{rebol2020frame, sibechi2019exploiting}, Conditional Random Field (CRF)~\cite{kundu2016feature}, non-local attention~\cite{hu2020temporally}. However, these methods either require accurate pixel-correspondence across frames, which is not always available or reliable, or utilize more complicated models (e.g., attention), which introduce significant computational overhead. Recently, Liu et al.~\cite{liu2020efficient} have proposed incorporating optical flow only during training and utilizing distillation to derive smaller networks, making inference more efficient. However, their method still requires video data for training and is constrained by the estimated optical flow's reliability. Some other works exploit the temporal nature of videos to improve accuracy~\cite{gadde2017semantic, li2021dynamic, pasad2020improving, wang2021temporal,chen2020naive,hu2020real} and not for improving consistency.

\vspace{0.01in}
\noindent \textbf{Test-Time Adaptation:} Past studies have also looked at test-time model updates, e.g., Tent~\cite{wang2020fully} for domain adaptation. Online adaptation could be another route for improving temporal consistency. Some works utilize optical flow during the test for adaptation, which requires multi-frame processing and considerable computations~\cite{bonneel2015blind, miksik2013efficient, yao2017occlusion}. Recently, Lei et al.~\cite{lei2020blind} proposed DVP for single-video test-time adaptation, which has been shown to work well for low-level vision tasks, e.g., colorization. However, DVP requires significant model update effort during test time. Some other works assume that the ground-truth annotations are available for the first frame and consider the adaptation as a few-shot learning problem~\cite{meinhardt2020make, voigtlaender2017online, xiao2019online}, which is a different setup than what we analyze in our paper.

\vspace{-0.0in}
\section{Proposed Method}\label{sec:method}
\vspace{-0.05in}

When a semantic segmentation network is uncertain about its estimates (i.e., the top two classes having numerically similar responses), small input variations could result in contrasting segmentation decisions due to the hard-decision rule (i.e., argmax) applied at the end of the network. An example of this issue can be seen in Fig.~\ref{fig:illustration_tc} (1st and 3rd columns), where similar road pixels are assigned to different labels across time. Such output uncertainty constitutes a major cause of temporal inconsistency in video segmentation. 

This section describes the proposed AuxAdapt method that improves temporal consistency by enabling the network to learn from its own hard decisions. Note that soft decisions do not generate a loss and thus are not useful in an unsupervised setting. Online adaptation encourages the network to be more confident in its outputs, leading to a more consistent semantic segmentation across frames. 

AuxAdapt employs an auxiliary network (AuxNet) to aid the adaptation. AuxNet is tiny; thus, it has a small computation overhead. Not updating the given main network (MainNet) improves the overall stability and prevents catastrophic forgetting. It also helps to maintain accuracy. Adapting only the AuxNet branch makes the process more flexible, allowing it to apply most segmentation architectures. In addition, AuxNet can utilize intermittent (conditional) updates and spatial sampling to reduce computations further. We also bring in an adaptive momentum scheme to decide how much of the previous information to use for adaptation, based on the difference between two consecutive frames.

\subsection{Learning from Network's Own Decisions}
\vspace{-0.05in}
Consider a pretrained semantic segmentation network $f^\text{main}$, 
which takes as input an RGB image, $x \in [0,\,1]^{H \times W \times 3}$, and generates a response/prediction map, ${y}^\text{main} \in \mathbb{R}^{H \times W \times K}$, for a $K$-class semantic segmentation task, where $H$ and $W$ are the height and width of the input image, respectively. Given a spatial location $(i,\,j)$ in the image, ${y}^\text{main}(i,\,j,\,k)$ is a score that indicates how likely this pixel belongs to class $k$. For the segmentation decision, an argmax operation is applied pixel-wise to the last dimension of ${y}^\text{main}$, such that the most probable class is assigned to each pixel. We denote this hard (discrete) decision as ${y}^\text{seg} \in \{1,2, \dots, K\}^{H \times W}$. To obtain semantic segmentation for a sequence of video frames $X=\{x_1,\, \dots,\, x_T\}$, $f^\text{main}$ can be applied to generate $Y=\{y^\text{seg}_1,\, \dots,\, y^\text{seg}_T\}$, where $T$ is the number of frames in the video. 

While applying $f^\text{main}$ to $X$ can generate video semantic segmentation, the resulting output $Y$ is usually temporally inconsistent, as shown in Fig.~\ref{fig:illustration_tc} (1st and 3rd columns). This is mainly due to the network's uncertainty in its own output, as aforementioned. To improve the temporal consistency, it is necessary to reduce the network's uncertainty. We address this by training the network on its own hard decisions, which reinforces its own belief. This allows the network to generate more confident predictions for image regions that are visually similar to what it has seen before. Our method works in a one-pass fashion, only requiring the network to go through the video once. The adaptation based on the current frame will immediately lead to more temporally consistent segmentation on the next frame, as consecutive frames share similar visual content. Note that existing test-time adaptation methods, e.g., DVP~\cite{lei2020blind}, Tent~\cite{wang2020fully}, do not take advantage of the sequential property (i.e., consecutive frames are visually similar) and need to re-apply the updated network to the full video again. DVP also requires training at least 25 epochs for adapting to a video. 

One possible, straightforward way to implement such an adaptation scheme is to compute the loss between $y^\text{main}_t$ and $y^\text{seg}_t$ and update the network on this loss, at each time $t$. We refer to this adaptation scheme as \textbf{NaiveAdapt}, which is illustrated in Fig.~\ref{fig:overview} (middle). This NaiveAdapt approach, however, has some drawbacks. First, performing backward pass on the entire network is computationally expensive.\footnote{An alternative is to update the last layer(s) of the network, however, this does not provide meaningful improvements, as we shall see in the experiments in Sec.~\ref{sec:exp_results}.} Also, adapting the segmentation network for a long video can degrade its general segmentation accuracy. Moreover, updating the network can be tricky for architectures that require information from previous frames, such as TDNet~\cite{hu2020temporally} and those with recurrent modules.

\SetInd{0.3em}{0.3em} 
\setlength{\textfloatsep}{5pt} 
\begin{algorithm}[t!]
\small
\SetAlgoLined
\textbf{Input:} $x_1,\, x_2,\, \dots,\, x_T$\;
\textbf{Output:} ${y^\text{seg}_1},\, {y^\text{seg}_2},\, \dots,\, {y^\text{seg}_T}$\;
Load trained MainNet $f^\text{main}$, which will be frozen\;
Load trained AuxNet $f^\text{aux}$ as $f^\text{aux}_1$\;
Initialize $t=1$\;
 \While{$t \leq T$}{
  $y^\text{main}_t = f^\text{main}(x_t)$,\footnotemark \ \
  $y^\text{aux}_t = f^\text{aux}_t(x_t)$\;
  $y^\text{seg}_t(i,\,j) \! =\! \underset{k}{\text{argmax}}\ y^\text{main}_t(i,\,j,\,k)+y^\text{aux}_t(i,\,j,\,k), \forall (i,\,j)$\;
Compute loss: $\mathcal{L}(y^\text{aux}_t, y^\text{seg}_t)$ using Eq.~(\ref{eq:loss});\\
Update $f^\text{aux}_t$ using Eq.~(\ref{eq:update_aux}), which gives $f^\text{aux}_{t+1}$\;
  $t \gets t +1 $\;
 }
 \caption{AuxAdapt}
 \label{alg:auxadapt}
\end{algorithm}
\footnotetext{While our algorithm description assumes that MainNet operates on each individual frame, AuxAdapt is compatible with those that utilize information from multiple frames (e.g.,~\cite{hu2020temporally}): $y^\text{main}_t = f^\text{main}(x_t,\, x_{t-1},\, ...,\, x_{1})$, as we shall see in Sec.~\ref{sec:exp_results}.}


\subsection{Adaptation Using Auxiliary Network}
\vspace{-0.05in}
In order to overcome the disadvantages of NaiveAdapt, we propose \textbf{AuxAdapt}, which employs a separate auxiliary network (AuxNet), $f^\text{aux}$, to work with the main segmentation network (MainNet), $f^\text{main}$, during the adaptation process. AuxNet is a separately-trained small-sized segmentation network. When streaming the video, at each time $t$, MainNet and AuxNet produce their respective prediction maps for the current frame, $y^\text{main}_t$ and $y^\text{aux}_t$. An argmax operation is then applied to the summation of these two maps to obtain the discrete semantic segmentation decision, $y^\text{seg}_t$. 
Unlike NaiveAdapt, in AuxAdapt, only the AuxNet is updated based on $y^\text{seg}_t$ while MainNet is kept frozen. AuxNet is updated using gradient descent as follows: 
\begin{equation}\label{eq:update_aux}
\begin{split}
    & \Delta \theta_{t}^\text{aux} = \beta \Delta \theta_{t-1}^\text{aux} + \alpha \nabla_{\theta^\text{aux}} \mathcal{L}(y^\text{aux}_t,\, y^\text{seg}_t),\\
    & \theta_t^\text{aux} = \theta_{t-1}^\text{aux} + \Delta \theta_t^\text{aux},
\end{split}
\end{equation}
where $\theta^\text{aux}$ denotes the parameters of AuxNet, $\alpha$ is the learning rate, $\beta$ is a momentum coefficient that controls the contribution of past gradients. $\mathcal{L}$ is the loss function: 
\begin{equation}\label{eq:loss}
    \mathcal{L}(y^\text{aux}_t,\, y^\text{seg}_t)= \sum_{i=1}^{H} \sum_{j=1}^{W} \frac{\mathcal{L}_\text{CE}(y^\text{aux}_t(i,\,j),\,y^\text{seg}_t(i,\,j))}{HW},
\end{equation}
where $\mathcal{L}_\text{CE}$ is the cross-entropy loss (with softmax)~\cite{CEL}. 


Since the updates are performed as the network consumes each frame of the video, the batch size is 1. As such, the mean and standard deviation of batch normalization layers are fixed during this process.

During this process, although MainNet is frozen, the adaptation is enabled by combining the output of the adaptable AuxNet and the output of the MainNet to generate the final discrete segmentation decision for each frame. Then, AuxNet learns from the discrete final decision to reduce the uncertainty in the overall prediction. Our AuxAdapt algorithm is summarized in Algorithm~\ref{alg:auxadapt}.

AuxAdapt offers several advantages. Efficiency-wise, by adopting a small AuxNet, the computation of its forward and backward passes will be much smaller than that of updating the entire MainNet. In addition, we can easily apply AuxAdapt to any semantic segmentation network, as we only need to run forward pass on MainNet and avoid involving the possibly intricate training procedures of MainNet.

\begin{figure}[t!]
    \centering
    \includegraphics[width=0.98\linewidth]{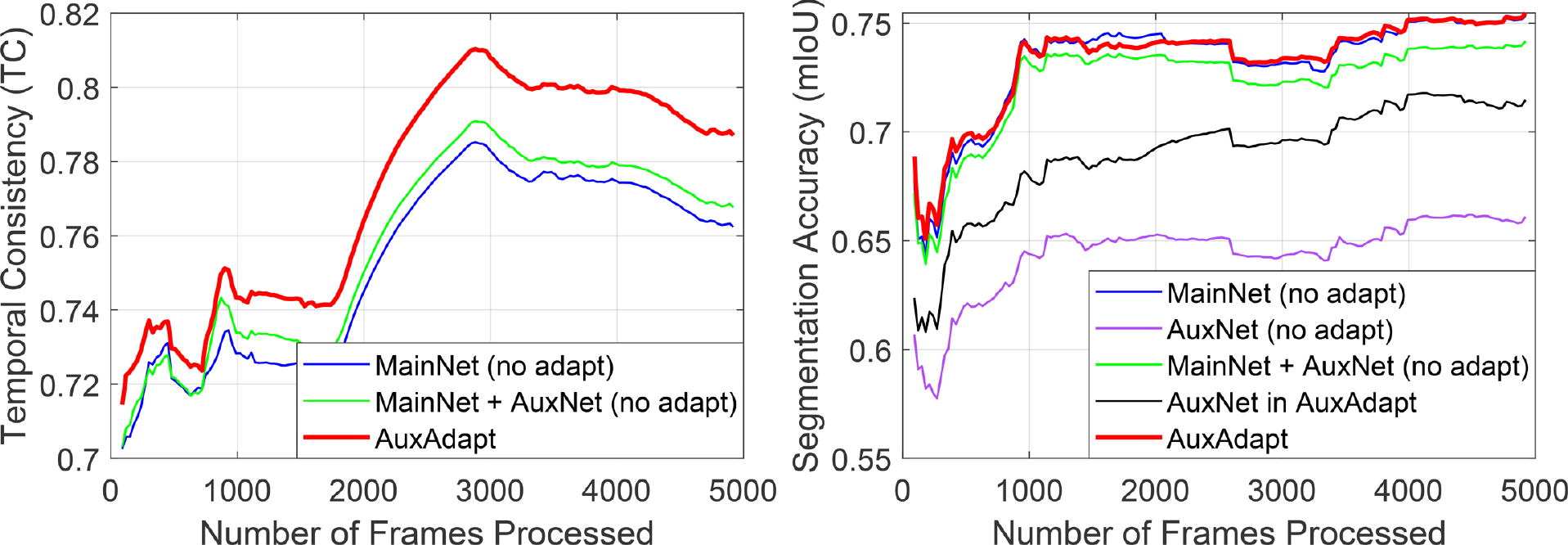}
    \caption{\small Adaptation dynamics of AuxAdapt on a CamVid test video. AuxAdapt improves temporal consistency (left) and maintains overall accuracy (right) in online adaptation.} 
\label{fig:miou_and_tc_dynamics}
\vspace{5pt}
\end{figure}

Furthermore, AuxAdapt can maintain the segmentation accuracy both on the given test video and on other general test images, thanks to our integrated MainNet-AuxNet model during the adaptation. More specifically, as MainNet is unchanged, we fundamentally prevent catastrophic forgetting. In contrast, updating MainNet itself (i.e., NaiveAdapt) can degrade segmentation accuracy, as we shall see in Sec.~\ref{sec:exp_results}. In addition, combining MainNet's and AuxNet's outputs for the overall segmentation has a number of benefits. First, the variation of accuracy is limited due to MainNet's contribution to the overall output. Learning from the aggregated decisions allows AuxNet to distill knowledge from MainNet, which encourages AuxNet to match MainNet's performance. Lastly, the ensemble effect of the two networks also helps maintain (sometimes even improve) the segmentation accuracy.

Figure~\ref{fig:miou_and_tc_dynamics} shows the adaptation dynamics for a sample test video. Fig.~\ref{fig:miou_and_tc_dynamics} (left) shows that AuxAdapt (red) considerably improves the temporal consistency of the original MainNet (blue) while streaming the video. During this process, it can be seen in Fig.~\ref{fig:miou_and_tc_dynamics} (right) that AuxAdapt (red) closely matches the accuracy of the original MainNet (blue). With AuxAdapt, AuxNet improves its accuracy by learning from MainNet during the process (black) as compared to the non-adapted AuxNet (purple). 

\begin{remark}
    While it is desirable for AuxNet to be very efficient, AuxNet should still provide a reasonable standalone segmentation accuracy. This can be achieved by utilizing a state-of-the-art efficient model, e.g., HRNet-w18~\cite{wang2020deep}. We note that it is not necessary for AuxNet to be as accurate as MainNet. In our experiments, we test various MainNet-AuxNet combinations where the AuxNet is usually a few percentage points less accurate than MainNet, but AuxAdapt is able to maintain the overall segmentation accuracy.
\end{remark}

\subsection{Reducing Computation Further}
\vspace{-0.06in}
In addition to utilizing an AuxNet, it is possible to further improve the efficiency of the adaptation process, by exploiting the redundancies in the temporal and spatial domains.

\vspace{0.00in}
\noindent \textbf{Intermittent Adaptation:} The consecutive frames in a video are usually highly similar, especially for a video with a large frame rate. As such, we do not need to update the network for every single frame and can instead perform the model update every several frames. As we shall show in our experiments, performing such intermittent adaptation further reduces computation while still providing a considerable improvement on temporal consistency.

\vspace{0.00in}
\noindent \textbf{Confidence-Based Spatial Sampling:} When the network is already highly confident in its prediction on a pixel, the loss (i.e., $\mathcal{L}_\text{CE}$ for pixel $(i,\,j)$) for this spatial location will be very small. Such loss terms will not provide meaningful contributions to updating the model and will instead incur unnecessary computation. As such, we can remove such redundant spatial locations from the overall loss computation by setting a confidence threshold. For each pixel, when the combined segmentation confidence (the highest softmax score among the $K$ classes) is above the given threshold, the corresponding loss will not be included in the model update.

\subsection{Motion-Based Adaptive Momentum}
\vspace{-0.06in}
How the network adapts should also depend on how fast the video scene evolves. For instance, given a slowly-moving scene, earlier frames can still be relevant and visually similar to the current frames. On the other hand, for a fast-changing scene, older frames can quickly become irrelevant. Momentum provides a mechanism for the network to control how much it should utilize previous information via the discounted gradients associated with earlier frames. A larger momentum (i.e., a larger $\beta$ in Eq.~(\ref{eq:update_aux})) allows the network to retain more past information and thus can benefit temporal consistency. This, however, might reduce segmentation accuracy as the network's update is mixed with the decisions on the earlier, less related frames. On the other hand, a smaller momentum makes the network focus more on the current frame and better preserves the original segmentation accuracy but provides less temporal consistency improvement.  To achieve a good balance, we propose using a motion-based adaptive scheme for the momentum coefficient. More specifically, we set $\beta = 1 - \frac{1}{HW}\|x_t - x_{t-1}\|_1$, where $\|\cdot \|_1$ denotes $L_1$ norm. This sets the strength of the momentum conditioned on how fast the frames change.

\section{Experiments}\label{sec:exp}
\vspace{-0.05in}
In this section (and also in Supplementary File), we present a comprehensive performance analysis on large benchmark datasets, compare with baselines and the current state of the art, assess the cross-dataset adaptation ability of our method, and evaluate our method under intermittent adaptation and spatial sampling, as well as our adaptive momentum scheme. Finally, we conduct in-depth ablation studies on various aspects of our method. 


\subsection{Experiment Setup}
\vspace{-0.05in}

\noindent \textbf{Datasets:} We run extensive performance evaluations of our method on Cityscapes (CS)~\cite{cordts2016cityscapes}, Camvid~\cite{brostow2009semantic}, and KITTI~\cite{geiger2013vision}. For CS, we use the validation set of 500 videos, each with 30 frames of size 1024$\times$2048. For CamVid, we use the test set, which contains two long video sequences with a frame size of 720$\times$960. To ensure a fair comparison with the latest video semantic segmentation method~\cite{liu2020efficient}, we use their setup for evaluating temporal consistency.

We use KITTI for cross-dataset adaptation experiments, i.e., the networks trained on CS are used for test-time adaptation on KITTI. To accommodate the class number and class label assignment difference between CS and KITTI, we map the 7 overlapping classes from KITTI to CS and treat all the other classes from both datasets as a separate class named ``others".  We utilize raw KITTI data for the test-time adaptation task, containing 60 videos, capped at 100 frames each. The frame size is 384$\times$1280. For evaluating segmentation accuracy, we use the 200 images that have ground-truth annotations, which are sampled across videos. 

\vspace{0.01in}
\noindent \textbf{Networks:} We use several state-of-the-art models as the main segmentation network (MainNet), including HRNet~\cite{wang2020deep}, DeepLabV3+~\cite{chen2018encoder} with a ResNet-101 backbone~\cite{he2016deep}, ETC~\cite{liu2020efficient} with a PSPNet-18 backbone~\cite{zhao2017pyramid}, and TDNet~\cite{hu2020temporally} with a PSPNet-18 backbone. For HRNet, we use two versions: HRNet-w48-s4 and HRNet-w18-s4.\footnote{The number following ``w" indicates the channel multiplier and the number following ``s" indicates the up-sampling ratio towards the output.} As for AuxNet, we use HRNet-w18-s8 and HRNet-w16-s8, which are light-weight models. The MainNet-AuxNet combinations are summarized in Table~\ref{tab:results_MainAux}.\footnote{When no trained AuxNets are available, we can create an AuxNet using a low-resolution copy of MainNet. This provides temporal consistency improvement comparable to that of a separately-trained AuxNet, but can incur slightly more computation as its architecture is dictated by MainNet. Detailed results can be found in the Supplementary File.}

\vspace{0.01in}
\noindent \textbf{Hyperparameters:} 
We use a learning rate of 0.0001 for all our experiments, a typical value for model fine-tuning.

\rowcolors{2}{white}{gray!09}
\begin{table}[t!]
\centering
\small
\begin{tabular}[h]{c |c | c }
\hline
\verydarkgray 
\textbf{Test Set} & \textbf{MainNet}   & \textbf{AuxNet}   	\\
\hline
CS, KITTI   &HRNet-w48-s4~\cite{wang2020deep}            & HRNet-w18-s8    \\
CS          &HRNet-w18-s4~\cite{wang2020deep}            & HRNet-w16-s8    \\
CS          &DeepLabV3+~\cite{chen2018encoder} (RN-101)  & HRNet-w18-s8    \\
CS          &ETC~\cite{liu2020efficient} (PSP-18) &HRNet-w18-s8 \\
CS          &TD4~\cite{hu2020temporally} (PSP-18) &HRNet-w16-s8 \\
CamVid      &HRNet-w18-s4~\cite{wang2020deep}             & HRNet-w16-s8    \\
CamVid      &PSPNet-101~\cite{zhao2017pyramid}               & HRNet-w18-s8   \\
CamVid      &WideResNet-38~\cite{zhu2019improving}               & HRNet-w18-s8   \\
\hline
\end{tabular}
\vspace{5pt}
\caption{\small MainNet-AuxNet combinations used in our main results shown in Tables~\ref{tab:results_cs},~\ref{tab:results_camvid},~and~\ref{tab:results_kitti}.}
\label{tab:results_MainAux}
\vspace{5pt}
\end{table}

\subsection{Evaluation Metrics}
\vspace{-0.05in}
To evaluate segmentation accuracy, we use the standard mean Intersection-over-Union (mIoU). For temporal consistency (TC) evaluations, we use FlowNet2~\cite{ilg2017flownet} to compute optical flow between two adjacent frames and warp the segmentation at frame $t$ to frame $t-1$. We then compare the warped and actual segmentations for each frame ($t<T$) using mIoU. The overall mIoU then serves as the TC metric. Note that this is the \textbf{same} TC metric used in~\cite{liu2020efficient}. We evaluate the computation efficiency of a method based on the MAC count in the forward and backward passes per frame. Notice, the backward pass costs twice as many MACs as the forward pass~\cite{BPcost}. 

\rowcolors{2}{white}{gray!09}
\begin{table}[t!]
\centering
\small
\begin{tabular}[h]{ c  | c | c | c }
\hline
\verydarkgray 
\textbf{Method}     & {TC}   &{mIoU}   &  {GMAC/$\!_F$}    \\ \hline
CC~\cite{shelhamer2016clockwork} & 71.2     & 67.7     &  - 	\\
DFF~\cite{zhu2017deep} & 71.4   & 68.7  &  - 	\\
Accel~\cite{jain2019accel} & 70.3 & 72.1 &  - 	\\
HRNet-w18-s8  \cite{wang2020deep}       & 71.9 &  72.6  &   19  \\
ETC~\cite{liu2020efficient} (PSP-101) & 71.7     & 79.5     &  1731 	\\
\hline
HRNet-w48-s4  \cite{wang2020deep}       & 72.1  &  81.0  &   750  \\
w/ DVP \cite{lei2020blind}   & 77.4 & 59.0 &   56924  \\
w/ Tent \cite{wang2020fully}  &  29.8   &   62.4   & 3000 \\
\textbf{w/ AuxAdapt (ours)}  & 75.8  &   {81.0}  &  808\\
\hline
HRNet-w18-s4 \cite{wang2020deep}       & 70.5 &  76.2  & 78    \\
w/ DVP \cite{lei2020blind}    &  76.9    &  73.6 &   5898 \\
w/ Tent \cite{wang2020fully}  &   55.3    &    61.3   & 310 \\
\textbf{w/ AuxAdapt (ours)}  & 75.3  &   {76.6}  & 128 \\

\hline
DeepLabV3+~\cite{chen2018encoder}		& 71.0     & 76.2   &  633 	\\
w/ DVP \cite{lei2020blind}		&   72.2    &   76.2    &   48108	\\
w/ Tent \cite{wang2020fully}		&     25.2  &   59.7    &   2532	\\
\textbf{w/ AuxAdapt (ours)}		&    75.1  &   76.6    &   691\\
\hline
ETC~\cite{liu2020efficient} (PSP-18)		& 70.6     & 73.1     &  463 	\\
w/ DVP \cite{lei2020blind}		&    76.2   &    71.5   &   35188	\\
w/ Tent \cite{wang2020fully}		&    57.1   &   58.6    &   1852	\\
\textbf{w/ AuxAdapt (ours)}		&    76.2   &   75.1    &   514	\\
\hline
TD4~\cite{hu2020temporally} (PSP-18) & 71.6     & 76.8      &  239 \\
\textbf{w/ AuxAdapt (ours)}		&    74.5   &   77.1    &   290	\\
\hline
	\end{tabular}
	\vspace{5pt}
	\caption{Performance evaluation on Cityscapes validation set.} 
	\label{tab:results_cs}
	\vspace{5pt}
\end{table}

\subsection{Results} \label{sec:exp_results}
\vspace{-0.05in}
 
We compare our proposed approach with various state-of-the-art (SOTA) methods, including 1) non-adaptive methods that train and test on single frames~\cite{chen2018encoder, jain2019accel, shelhamer2016clockwork, wang2020deep, zhao2017pyramid, zhu2017deep, zhu2019improving}, 2) non-adaptive methods that utilize multi-frame information (e.g., optical flow, temporal attention)~\cite{hu2020temporally, liu2020efficient}, and 3) adaptation methods~\cite{lei2020blind,wang2020fully}, as well as baseline adaptation methods that update MainNet without using AuxNet. Finally, we present results using our proposed motion-based adaptive momentum. 

\vspace{0.01in}
\noindent\textbf{Evaluation on Cityscapes \& Camvid:}
On Cityscapes, as summarized in Table~\ref{tab:results_cs}, AuxAdapt significantly improves the temporal consistency of the state-of-the-art models while preserving (and in some cases, improving) the segmentation accuracy. For instance, for MainNet of HRNet-w48-s4, AuxAdapt improves the TC from 72.1 to 75.8, requiring only 7\% additional computation. Furthermore, our approach can also improve models that have already utilized multi-frame information, e.g., ETC~\cite{liu2020efficient} that utilizes optical flow during training, TDNet~\cite{hu2020temporally} that uses temporal attention, as shown in the last two blocks of Table~\ref{tab:results_cs}.

We see that the SOTA test-time adaptation methods do not provide reliable performance and cost much more computation. For instance, DVP~\cite{lei2020blind} incurs an unfavorable trade-off between temporal consistency and accuracy. It trains, from scratch, a new network to mimic the original network and requires early stopping to prevent overfitting to temporally inconsistent patterns. However, stopping too early leads to a large performance gap between the original and new models. As such, when DVP delivers a TC similar to ours, it considerably reduces segmentation accuracy, and when it attains the original accuracy, its TC improvement is minimal. Moreover, as DVP requires training a new network for 25 epochs (default setting), its computation is prohibitively high for test time (46--70$\times$ higher than AuxAdapt). As for Tent~\cite{wang2020fully}, it significantly degrades both TC and accuracy. Tent uses entropy minimization to update the batch normalization (BN) layers of the original network, which severely limits the adaptation capability. Furthermore, to update the BN layers, it is required to propagate through the entire network.


\rowcolors{2}{white}{gray!09}
\begin{table}[t!]
\centering
\small
\begin{tabular}[h]{ c | c | c | c }
\hline
\verydarkgray 
\textbf{Method}   & {TC}   &{mIoU}   &  {GMAC/$\!_F$}   	\\
\hline
DFF \cite{zhu2017deep}    &  78.0     &  66.0  &   -- \\
Accel ~\cite{jain2019accel}  &  76.2     &  66.7  &   -- \\
\hline
HRNet-w18-s4 \cite{wang2020deep}    &   75.8     &  73.2   &  26  \\
w/ DVP~\cite{lei2020blind}&  56.6 &  71.4    &  1946\\
w/ Tent~\cite{wang2020fully}	      &  59.1 &  29.2    &  102\\ 
\textbf{w/ AuxAdapt  (ours)} 	      &  79.1 &  73.2    &  42\\
\hline
PSPNet-101 \cite{zhao2017pyramid}   &   76.7     &  76.2   &  691 \\
w/ DVP~\cite{lei2020blind}&  51.3 &  72.5    &  52546\\
w/ Tent~\cite{wang2020fully}	      &  20.6 & 28.3     & 2765\\ 
\textbf{w/ AuxAdapt (ours)} 	      &  79.5 &  76.4    &  711\\ \hline
WideResNet-38 \cite{zhu2019improving}    &   78.1    &  80.6   &  1920  \\
\textbf{w/ AuxAdapt (ours)} 	      & 79.4  &  80.8    &  1995\\
\hline
\end{tabular}
\vspace{5pt}
\caption{\small Performance evaluation on Camvid test set.}
\label{tab:results_camvid}
\vspace{-0pt}
\end{table}

\rowcolors{2}{white}{gray!09}
\begin{table}[t!]
\centering
\small
\begin{tabular}[h]{ c | c | c | c }
\hline
\verydarkgray 
\textbf{Method}   & {TC}   &{mIoU}   &  {GMAC/$\!_F$}   	\\
\hline
HRNet-w48-s4~\cite{wang2020deep} &57.4   &65.9   &   176  \\
DVP~\cite{lei2020blind}       &64.1  &27.2    &13361\\
\textbf{w/ AuxAdapt (ours)}	      & 63.5   & 65.8   & 189 \\
\hline
\end{tabular}
\vspace{5pt}
\caption{\small Cross-dataset adaptation from Cityscapes to KITTI.}
\label{tab:results_kitti}
\vspace{5pt}
\end{table}

On CamVid, as shown in Table~\ref{tab:results_camvid}, AuxAdapt consistently improves temporal consistency and maintains the segmentation accuracy for a longer test video. In contrast, DVP degrades in both TC and accuracy, as the default 25 epochs are not sufficient for the larger test data, and there is no guideline on how to increase the training time w.r.t. the video length properly. Similarly, Tent does not perform well as the much longer test video makes it challenging for training. Note that CamVid videos are much longer (6000+ frames). 

\vspace{0.01in}
\noindent\textbf{Cross-Dataset Adaptation:}
In practice, the test data can have different characteristics than the training data, e.g., scenes and camera settings. Here, we evaluate AuxAdapt for such a scenario by adapting a Cityscapes-trained model to KITTI data. Note that the segmentation accuracy is evaluated on KITTI based on the 200 images (from different videos!) with ground-truth annotations. Such an evaluation will assess whether the adapted model maintains a general segmentation capability beyond the one specific test video.

The results are summarized in Table~\ref{tab:results_kitti}. First, it can be seen that the performance of MainNet (without adaptation) drops considerably as compared to that on Cityscapes, due to the difference between KITTI and Cityscapes. By applying AuxAdapt, the TC significantly improves from 57.4 to 63.5. 
We also compare with DVP. While DVP improves temporal consistency, it incurs significantly heavier computation, and the obtained model obtained cannot provide acceptable segmentation accuracy on images outside the given test video. In other words, the DVP-trained model cannot be applied to a new test video without incurring another expensive training session. On the other hand, AuxAdapt can readily provide accurate segmentation on new test data, without the need of further adapting.

\rowcolors{2}{white}{gray!09}
\begin{table}[t!]
\centering
\small
\begin{tabular}[h]{ c | c | c | c }
\verydarkgray 
\hline
\textbf{Method}     & {TC}   &{mIoU}   &  {GMAC/$\!_F$}\\
\hline
\darkgray \multicolumn{4}{c}{Cityscapes}\\ \hline
HRNet-w48-s4  & 72.1  &  81.0  &   750  \\
w/ NaiveAdapt (Last Part)  & 72.3 &   80.5  &  889\\
w/ NaiveAdapt (All Layers)    & 76.2    & 80.9    &   2249	\\
+ AuxNet (HRNet-w18-s8)    &  73.3 &    80.9  &  768\\
\textbf{w/ AuxAdapt (ours)}  & 75.8  &   {81.0}  &  808\\
\hline
HRNet-w18-s4  & 70.5 &  76.2  & 78\\ 
w/ NaiveAdapt (Last Part)  & 70.7 &   76.2  &  109\\ 
w/ NaiveAdapt (All Layers)   & 74.1  &  76.6   &  233\\ 
{+ AuxNet (HRNet-w16-s8)} & {72.9}  & 76.5    & 95\\ 
\textbf{w/ AuxAdapt (ours)} & 75.3  &   {76.6}  & 128\\ 
\hline



\darkgray \multicolumn{4}{c}{KITTI}\\ \hline
HRNet-w48-s4              &57.4   &65.9   &   176  \\
w/ NaiveAdapt (Last Part)	          & 57.3   & 65.7   &  208 \\
w/ NaiveAdapt (All Layers)	          & 62.3   & 62.5   &   527\\
+ AuxNet (HRNet-w18-s8)  &58.8   &65.2  & 180 \\
\textbf{w/ AuxAdapt (ours)}	      & 63.5   & 65.8   & 189  \\
\hline
	\end{tabular}
	\vspace{5pt}
	\caption{Comparison with baseline adaptation schemes, including two options of NaiveAdapt which updates the last part and all layers of MainNet, respectively (without AuxNet). We also report the performance of the integrated MainNet-AuxNet model, without model update (4th row of each block).} 
	\label{tab:results_mainadapt_vs_auxadapt}
	\vspace{5pt}
\end{table}

\vspace{0.01in}
\noindent\textbf{Comparing with Baseline Adaptation Schemes:}\label{sec:exp_baseline}
We compare AuxAdapt with the NaiveAdapt baseline on Cityscapes and KITTI. For NaiveAdapt, we consider two options: 1) updating only the last part,\footnote{The last part of HRNet consists of two conv. layers and a BN layer.} and 2) updating all the layers of MainNet. As shown in Table~\ref{tab:results_mainadapt_vs_auxadapt}, only updating the last part of MainNet does not result in meaningful TC gain, while updating the entire MainNet produces less or similar TC gain but incurs much higher computation. Note that for cross-dataset adaptation on KITTI, updating the entire MainNet considerably reduces accuracy due to overfitting to the given test video. As for the case of adding MainNet's and AuxNet's outputs without adaptation, TC increases slightly but is considerably lower than AuxAdapt, indicating that adaptation is key to improving TC.

\vspace{0.01in}
\noindent\textbf{Intermittent Adaptation:}
As shown in Table~\ref{tab:results_intermittent_update}, under intermittent adaptation, AuxAdapt still considerably improves TC and maintains the original accuracy, using even lower computation. Note that on KITTI, as the update happens less frequently, TC improvement drops slightly faster as compared to on Cityscapes. This is because the networks are more uncertain on KITTI as they are trained on Cityscapes and the scenes in KITTI evolve more quickly. In spite of these challenging factors, AuxAdapt is still able to considerably improve TC with intermittent adaptation.



\rowcolors{2}{white}{gray!09}
\begin{table}[t!]
\centering
\small
\begin{tabular}[h]{ c | c | c | c  }
\hline
\verydarkgray 
\textbf{Update Frequency}   & {TC}   &{mIoU}   &  {GMAC/$\!_F$}   	\\\hline
\darkgray \multicolumn{4}{c}{Cityscapes}\\
\hline
No Adaptation            & 72.1   & 81.0 & 750\\
Every Frame         	     & 75.8   & 81.0   & 808  \\
Every 2 Frames         	 & 75.4   & 81.0   & 789  \\
Every 5 Frames         	 & 74.8   & 81.0   &  777 \\
Every 10 Frames            & 74.7   & 80.9   &  773 \\
\hline
\darkgray \multicolumn{4}{c}{KITTI}\\
\hline
No Adaptation            & 57.4   & 65.9 & 176\\
Every Frame         	 & 63.5   & 65.8  & 189\\
Every 2 Frames         	  & 62.3   & 66.2 & 185\\
Every 5 Frames         	  & 60.6   & 66.3 & 182\\
Every 10 Frames           & 59.9   & 66.1  & 181\\
\hline
\end{tabular}
\vspace{0.05in}
\caption{\small Intermittent Adaptation. MainNet is HRNet-w48-s4 and AuxNet is HRNet-w18-s8.}
\label{tab:results_intermittent_update}
\vspace{-3pt}
\end{table}


\rowcolors{2}{white}{gray!09}
\begin{table}[t!]
\centering
\small
\begin{tabular}[h]{ c | c | c | c | c | c | c }
\hline
\verydarkgray 
  \textbf{Selection} & \multicolumn{3}{c|}{\textbf{Cityscapes}}   &\multicolumn{3}{c}{\textbf{KITTI}}\\ 
\verydarkgray \textbf{Criterion} & {TC}   &{mIoU} & \% &  {TC} &  {mIoU}& \% \\ \hline
None     & 72.1   & 81.0   &0    & 57.4   & 65.9  &0\\
All      & 75.8   & 81.0   &100  & 63.5   & 65.8  &100\\
Conf.$<$0.9  & 76.6   & 81.0   &15.0 & 63.8   & 64.1  &18.0\\
Conf.$<$0.8  & 76.5   & 81.0   &11.4 & 64.0   & 63.8  &13.9\\
\hline
\end{tabular}
\vspace{5pt}
\caption{\small Confidence-based spatial sampling. MainNet is HRNet-w48-s4 and AuxNet is HRNet-w18-s8. The 3rd column of each dataset block shows the average percentage of pixel locations included in the model updates. Note that ``None" is MainNet only (without adaptation) and ``All" is AuxAdapt without sub-sampling.}
\label{tab:results_spatial_sampling}
\vspace{5pt}
\end{table}

\vspace{0.01in}
\noindent\textbf{Confidence-Based Spatial Sampling:}
As shown in Table~\ref{tab:results_spatial_sampling}, by using confidence-based sampling, we significantly reduce the number of pixel locations included in the loss computation, which further enhances efficiency. Meanwhile, we achieve similar TC improvements. On KITTI, we notice a minor drop in mIoU when applying the spatial sampling. This is because the Cityscapes-trained networks are more uncertain on KITTI and excluding the confident pixels can limit AuxNet's learning from MainNet.



\begin{figure}[t!]
    \vspace{-0.13in}
    \centering
    \includegraphics[width=0.92\linewidth]{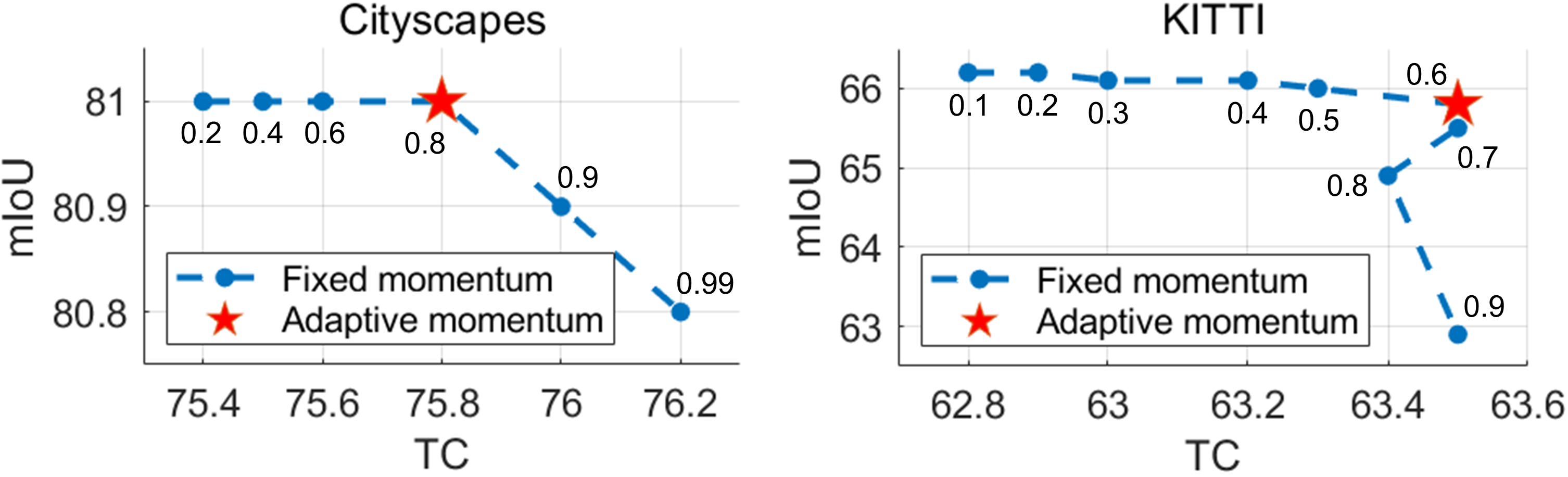}
    \vspace{-0.03in}
    \caption{TC and mIoU by using our motion-adaptive momentum coefficient, as compared to fixed momentum coefficients.} 
\label{fig:momentum}
\vspace{0.08in}
\end{figure}

\vspace{0.01in}
\noindent\textbf{Motion-Based Adaptive Momentum:}
In Figure~\ref{fig:momentum}, we show that our motion-adaptive momentum automatically finds a good balance between TC and accuracy across datasets, as compared to using a fixed momentum.

\vspace{-0.064in}
\section{Conclusions}\label{sec:conclude}
\vspace{-0.07in}
In this paper, we proposed a novel, unsupervised online adaptation method, AuxAdapt, for improving temporal consistency of video semantic segmentation in test time. AuxAdapt does not require optical flow (or cross-frame features) and can be applied to any segmentation network. By employing a small auxiliary network to work with the original segmentation network, AuxAdapt considerably boosts the temporal consistency of state-of-the-art segmentation models while using a tiny fraction of the computation required by existing adaptation methods. Furthermore, AuxAdapt exploits temporal and spatial sub-sampling for further acceleration, and utilizes an adaptive scheme to automatically set the momentum for adaptation.

{\small
\bibliographystyle{ieee_fullname}
\bibliography{egbib}
}

\clearpage
\section{Supplementary Materials}

\subsection{Per-Class Temporal Consistency}
As shown in Table~\ref{tab:classwise tc}, AuxAdapt provides consistent TC improvement across classes.

\renewcommand{\tabcolsep}{1pt}
\rowcolors{2}{white}{gray!09}
\begin{table}[h!]
\vspace{-8pt}
\centering
\footnotesize
\begin{tabular}[t!]{ c | c | c | c |c | c| c | c | c |c | c}
\hline
\verydarkgray 
\textbf{Method}   & {road}   &{sdwk.}   &  {bldg.}  & {wall}  &{fence} & {pole}   &{t-light}   &  {t-sign}  & {veg.} &{terrain}
\\\hline
No Adapt 
& 96.5 & 82.8 & 87.6 & 53.6 & 66.3 & 30.2 & 57.8 
& 58.5 & 88.9 & 76.5 \\
\hline      
AuxAdapt 
& 97.3 & 85.4 & 89.8 & 65.2 & 75.8 & 30.2 & 60.2
& 61.4 & 90.5 & 81.2 \\
\hline
\hline
\verydarkgray 
\textbf{Method}  &  {sky}  & {person}  &{rider}&  {car}  & {truck}  &{bus}	&{train} &{m-bike} &{bike} &\textbf{mean}\\\hline
No Adapt 
& 91.9 & 60.6 & 58.7 & 85.7 
& 81.1 & 85.9 & 83.0 & 56.1 & 68.1 & 72.1\\
 \hline      
AuxAdapt 
& 92.6 & 62.7 & 62.0 & 87.5 
& 85.7 & 90.4 & 86.7 & 65.1 & 71.3 & 75.8\\
\hline
\end{tabular}
\vspace{2pt}
\caption{Per-class temporal consistency on Cityscapes. MainNet: HRNet-w48-s4. AuxNet: HRNet-w18-s8.}
\label{tab:classwise tc}
\vspace{-5pt}
\end{table}

\subsection{When Pretrained AuxNet is Unavailable}
During test time, there can be cases where only the main segmentation network (MainNet) is provided and no pretrained auxiliary networks (AuxNets) are available. In such cases, it is still possible to apply AuxAdapt to improve temporal consistency, by creating a lower-resolution copy of MainNet to serve as AuxNet. More specifically, we create a copy of MainNet and add a down-sampling layer at the beginning of it. In this way, AuxNet shares the same architecture and weights as MainNet, but works with down-sampled inputs, thus considerably saving computation. At the end of AuxNet, a corresponding up-sampling layer is added such that the size of its output matches that of MainNet.

In Table~\ref{tab:results_ofm}, it can be seen that in this setting where AuxNet is obtained from MainNet (denoted as ``OFM"), AuxAdapt provides considerable improvement to temporal consistency while maintaining segmentation accuracy. As compared to the case where a pretrained AuxNet is available (denoted as ``PT"), AuxAdapt in the OFM setting provides very similar performance. Overall, the OFM option makes AuxAdapt more widely applicable while providing comparable adaptation performance. Note that, in this OFM setup, the additional computational cost for test-time adaptation could be higher than using a well-designed and pretrained AuxNet since now AuxNet is directly derived from MainNet, and is not optimized for its efficiency.


\rowcolors{2}{white}{gray!09}
\begin{table}[h!]
\centering
\small
\begin{tabular}[h]{ c | c | c | c }
\hline
\verydarkgray 
\textbf{Method}   & {TC}   &{mIoU}   &  {GMAC/$\!_F$}   	\\
\hline
\darkgray \multicolumn{4}{c}{Cityscapes}\\
\hline
HRNet-w18-s4 \cite{wang2020deep}   & 70.5 &  76.2  & 78\\ 
\textbf{w/ AuxAdapt (PT)}  &  75.3 & 76.6    & 128\\
\textbf{w/ AuxAdapt (OFM)} 	      &  75.2 & 76.4    & 136\\ 
\hline
\darkgray \multicolumn{4}{c}{CamVid}\\
\hline
HRNet-w18-s4 \cite{wang2020deep}    &   75.8     &  73.2   &  26   \\
\textbf{w/ AuxAdapt (PT)} 	      &  79.1 &  73.2    &  42\\ 
\textbf{w/ AuxAdapt (OFM)} 	      &  78.9 &  73.2    &  45\\ 
\hline
WRN38 \cite{zhu2019improving}    &   78.1    &  80.6   &  1920   \\
\textbf{w/ AuxAdapt (PT)} 	      & 79.4  &  80.8    &  1995\\
\textbf{w/ AuxAdapt (OFM)} 	      & 79.7  &  80.7    &  2280\\ 
\hline
	\end{tabular}
	\vspace{5pt}
	\caption{\small AuxAdapt using MainNet-derived AuxNet on Cityscapes and CamVid. OFM indicates that AuxNet is obtained from MainNet, with an additional down-sampling operation at the beginning. PT denotes the setting where AuxNet is pretrained using the corresponding architectures described in Table~1 of the main paper.}
	\label{tab:results_ofm}
	\vspace{5pt}
\end{table}

\subsection{Input Down-sampling for AuxNet}
In the main paper, $2\times$ down-sampling is applied to AuxNet's input (see Fig.~2 of main paper), which reduces computation. In this part, we study the effect of further down-sampling the input to AuxNet. 

Table~\ref{tab:results_kitti_input_resolution} shows the results with a more aggressive down-sampling ratio ($3\times$). It can be seen that  the TC improvement is similar and the computation cost is reduced. However, the segmentation accuracy slightly drops, as the further-down-sampled input now contains less information.

\rowcolors{2}{white}{gray!09}
\begin{table}[h!]
\centering
\small
\begin{tabular}[h]{ c | c | c | c }
\hline
\verydarkgray 
\textbf{Method}   & {TC}   &{mIoU}   &  {GMAC/$\!_F$}   	\\\hline
\darkgray \multicolumn{4}{c}{Cityscapes}\\
\hline
HRNet-w48-s4 \cite{wang2020deep}                                 & 72.1   &81.0   & 749.9  \\
w/ AuxAdapt ($2\times \downarrow$) & 75.8   &81.0   &  808.2\\
w/ AuxAdapt ($3\times \downarrow$) & 76.7   &80.5    &  776.2\\
\hline
\darkgray \multicolumn{4}{c}{KITTI}\\ \hline
HRNet-w48-s4 \cite{wang2020deep}             &57.4   &65.9   &   175.8  \\
AuxAdapt ($2\times \downarrow$) & 63.5   & 65.8   &  189.4\\
AuxAdapt ($3\times \downarrow$) & 63.4   & 64.0   &  181.9\\
\hline
\end{tabular}
\vspace{5pt}
\caption{\small Effect of AuxNet's input resolution. The numbers in the parentheses indicate how much the input image is down-sampled via average pooling. MainNet is HRNet-w48-s4. For $2\times$ ($3\times$) down-sampling, AuxNet is HRNet-w18-s8 (HRNet-w18-s12), where the number following ``s" indicates the upsampling ratio at the output.}
\label{tab:results_kitti_input_resolution}
\vspace{5pt}
\end{table}

\subsection{Standalone Performance of AuxNet}
In Table~\ref{tab:small_networks}, we report the \textbf{standalone} performance of the lightweight models which are used as AuxNets in our experiments. The TC, mIoU, and GMAC numbers reported here are based on using the AuxNet model alone, without MainNet and adaptation.

\rowcolors{2}{white}{gray!09}
\begin{table}[h!]
\centering
\small
\begin{tabular}[h]{ c | c | c | c}
\hline
\verydarkgray 
\textbf{Networks}   & {TC}   &{mIoU}   &  {GMAC/$\!_F$} \\ \hline 
\darkgray \multicolumn{4}{c}{Cityscapes}\\
\hline
HRNet-w18-s8 & 71.9    & 72.6    & 19\\  
HRNet-w16-s8  & 71.4   & 74.3   & 17\\   
\hline
\darkgray \multicolumn{4}{c}{CamVid}\\ \hline
HRNet-w18-s8             &76.8   &69.4 &6.3\\ 
HRNet-w16-s8             &76.5   &70.8 & 5.6\\  
\darkgray \multicolumn{4}{c}{KITTI}\\ \hline
HRNet-w18-s8             &54.1   &57.5  &4.3  \\ 
\hline
\end{tabular}
\vspace{5pt}
\caption{\small Standalone performance of the lightweight AuxNet models on Cityscapes, CamVid, and KITTI.}
\label{tab:small_networks}
\vspace{5pt}
\end{table}

\subsection{Output Uncertainty and Temporal Inconsistency}
Output uncertainty is a major cause of temporal inconsistency in video semantic segmentation. In this part, we visualize the uncertainty map of the segmentation output and analyze its connection to temporally inconsistent segmentation.

For each pixel $(i,\,j)$, we refer to the maximum value of the $K$-dimensional output of the softmax operation before applying argmax as the prediction confidence at this pixel, $c(i,\,j)$, where $K$ is the number of classes. For each pixel, we calculate the uncertainty as follows: $u(i,\,j) = 1 - c(i,\,j)$. We can then use $u$ as the uncertainty map of the network's segmentation decision, where for each pixel, a higher (lower) value indicates a higher (lower) uncertainty. 

\begin{figure}[t!]
    \centering
    \includegraphics[width=1.0\linewidth]{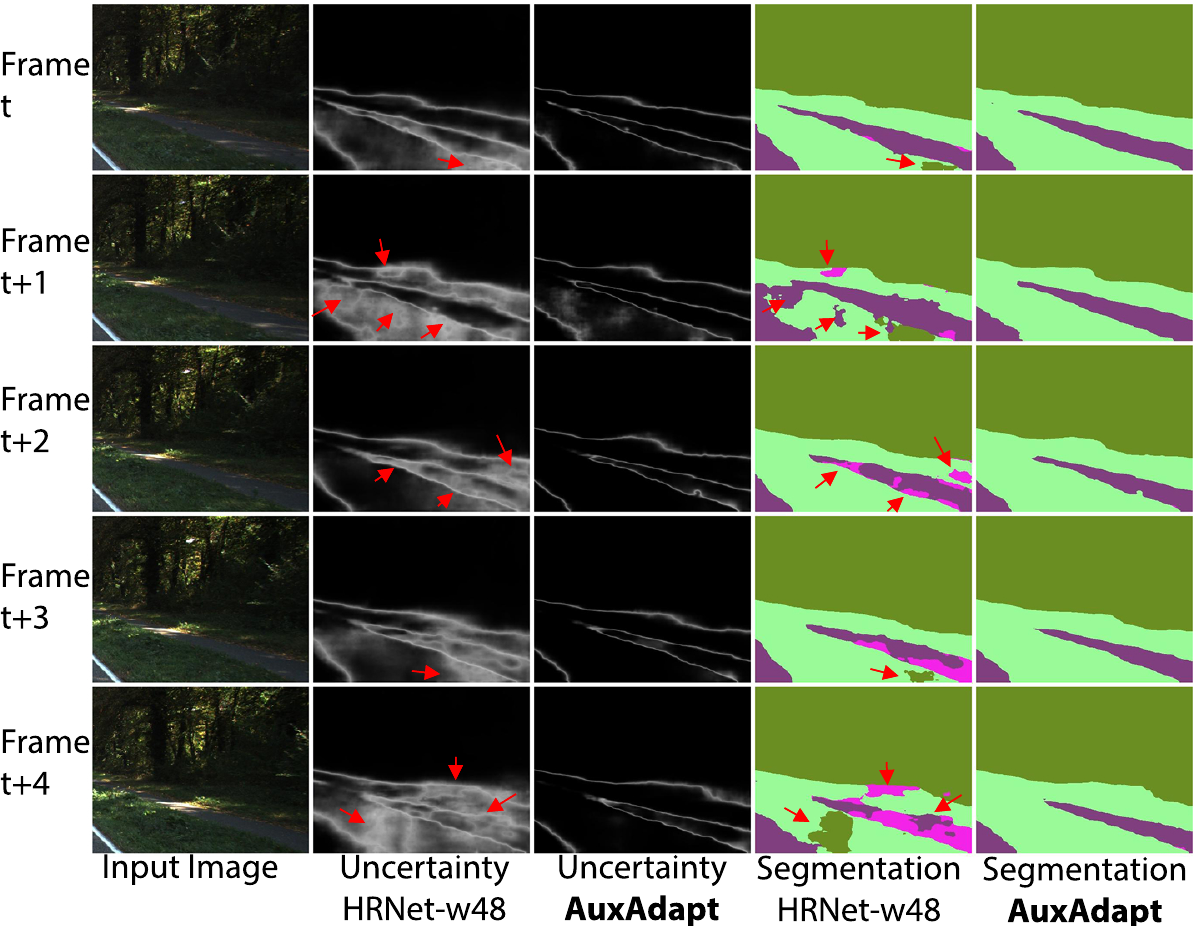}
    \caption{\small Connection between uncertainty and temporally inconsistency. Column~1 shows the input RGB frames. Columns~2~and~4 show the uncertainty maps and the segmentation decisions by using an HRNet-w48 model without adaptation. Columns~3~and~5 show the uncertainty maps and the segmentation decisions by using our proposed AuxAdapt. Red arrows indicate sample uncertain areas which lead to temporally inconsistent segmentations. It can be seen that AuxAdapt significantly reduces the output uncertainty and improves temporal consistency of the segmentation.} 
\label{fig:uncertainty_consistency1}
\vspace{0pt}
\end{figure}

\begin{figure}[]
    \centering
    \includegraphics[width=1.0\linewidth]{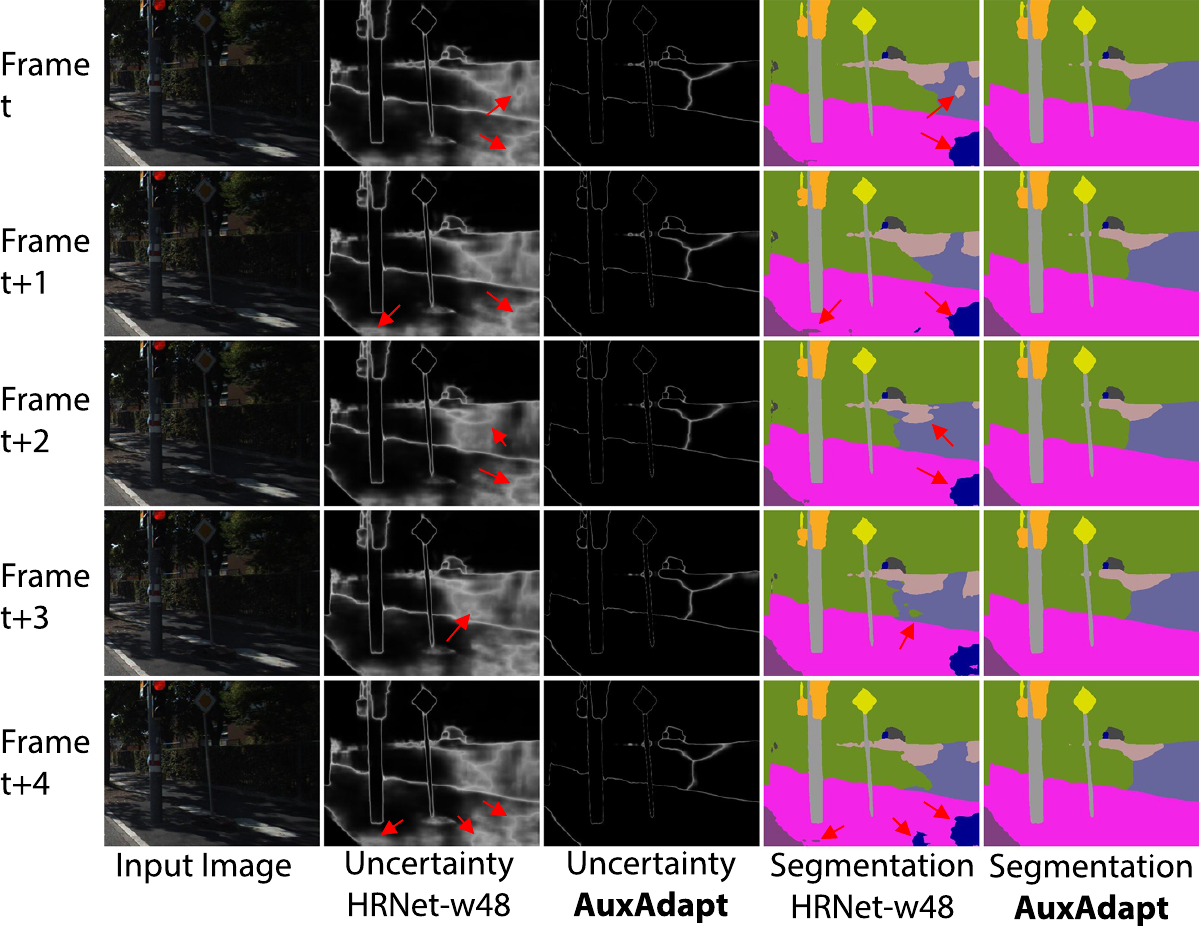}
    \caption{\small Connection between uncertainty and temporally inconsistency. Column~1 shows the input RGB frames. Columns~2~and~4 show the uncertainty maps and the segmentation decisions by using an HRNet-w48 model without adaptation. Columns~3~and~5 show the uncertainty maps and the segmentation decisions by using our proposed AuxAdapt. Red arrows indicate sample uncertain areas which lead to temporally inconsistent segmentations. It can be seen that AuxAdapt significantly reduces the output uncertainty and improves temporal consistency of the segmentation.} 
\label{fig:uncertainty_consistency2}
\vspace{0pt}
\end{figure}

\begin{figure}[]
    \centering
    \includegraphics[width=1.0\linewidth]{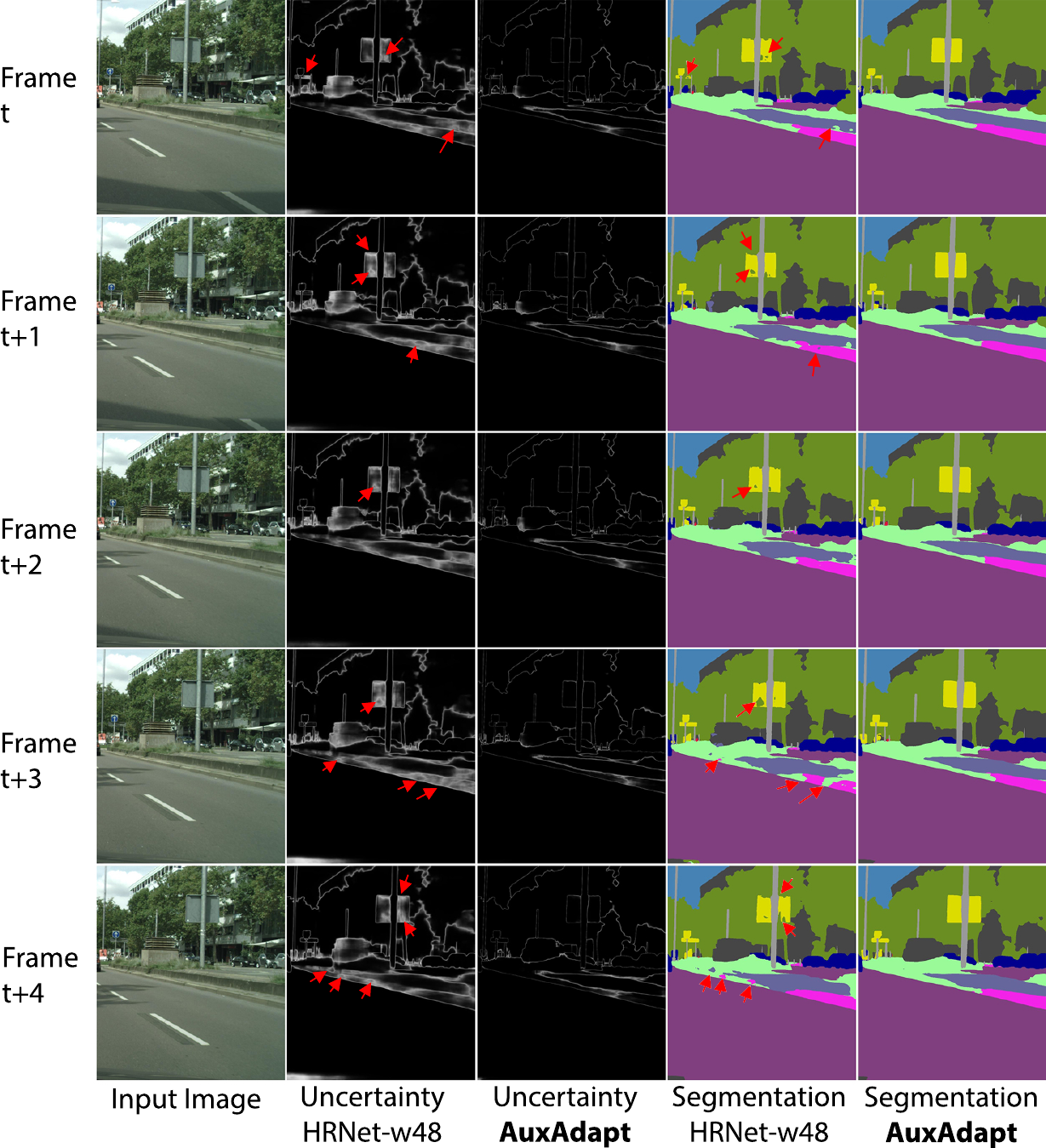}
    \caption{\small Connection between uncertainty and temporally inconsistency. Column~1 shows the input RGB frames. Columns~2~and~4 show the uncertainty maps and the segmentation decisions by using an HRNet-w48 model without adaptation. Columns~3~and~5 show the uncertainty maps and the segmentation decisions by using our proposed AuxAdapt. Red arrows indicate sample uncertain areas which lead to temporally inconsistent segmentations. It can be seen that AuxAdapt significantly reduces the output uncertainty and improves temporal consistency of the segmentation.} 
\label{fig:uncertainty_consistency3}
\vspace{0pt}
\end{figure}

In Fig.~\ref{fig:uncertainty_consistency1},~\ref{fig:uncertainty_consistency2}, and~\ref{fig:uncertainty_consistency3}, it can be seen that for the original model (MainNet), the uncertain regions (column~2) result in temporally inconsistent segmentation decisions (column~4). The red arrows indicate sample locations where uncertainties cause temporally inconsistent artifacts. On the other hand, AuxAdapt provides much more confident predictions (column~3), leading to significantly more temporally consistent results (column~5). 

\subsection{Comparing with State of the Art}
\begin{figure}[t]
    \centering
    \includegraphics[width=1.0\linewidth]{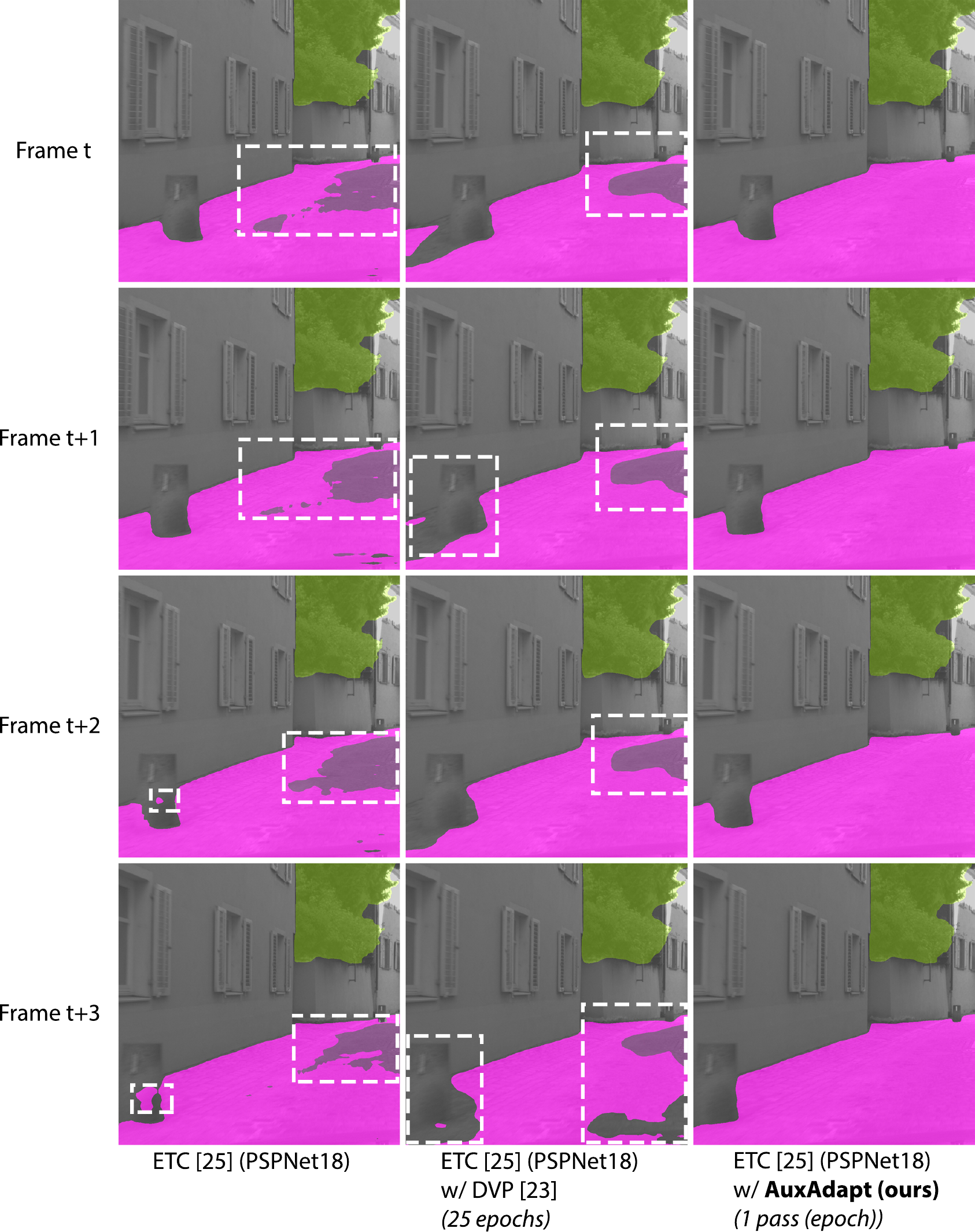}
    \caption{\small Visual results comparing our proposed AuxAdapt with state-of-the-art methods. \textbf{Left:} ETC~\cite{liu2020efficient}. \textbf{Middle:} DVP~\cite{lei2020blind}. \textbf{Right:} AuxAdapt (ours). White dashed boxes highlight temporally inconsistent and inaccurate segmentation. It can be seen that AuxAdapt significantly improves segmentation temporal consistency.} 
\label{fig:sample1}
\vspace{0pt}
\end{figure}

\begin{figure}[t!]
    \centering
    \includegraphics[width=1.0\linewidth]{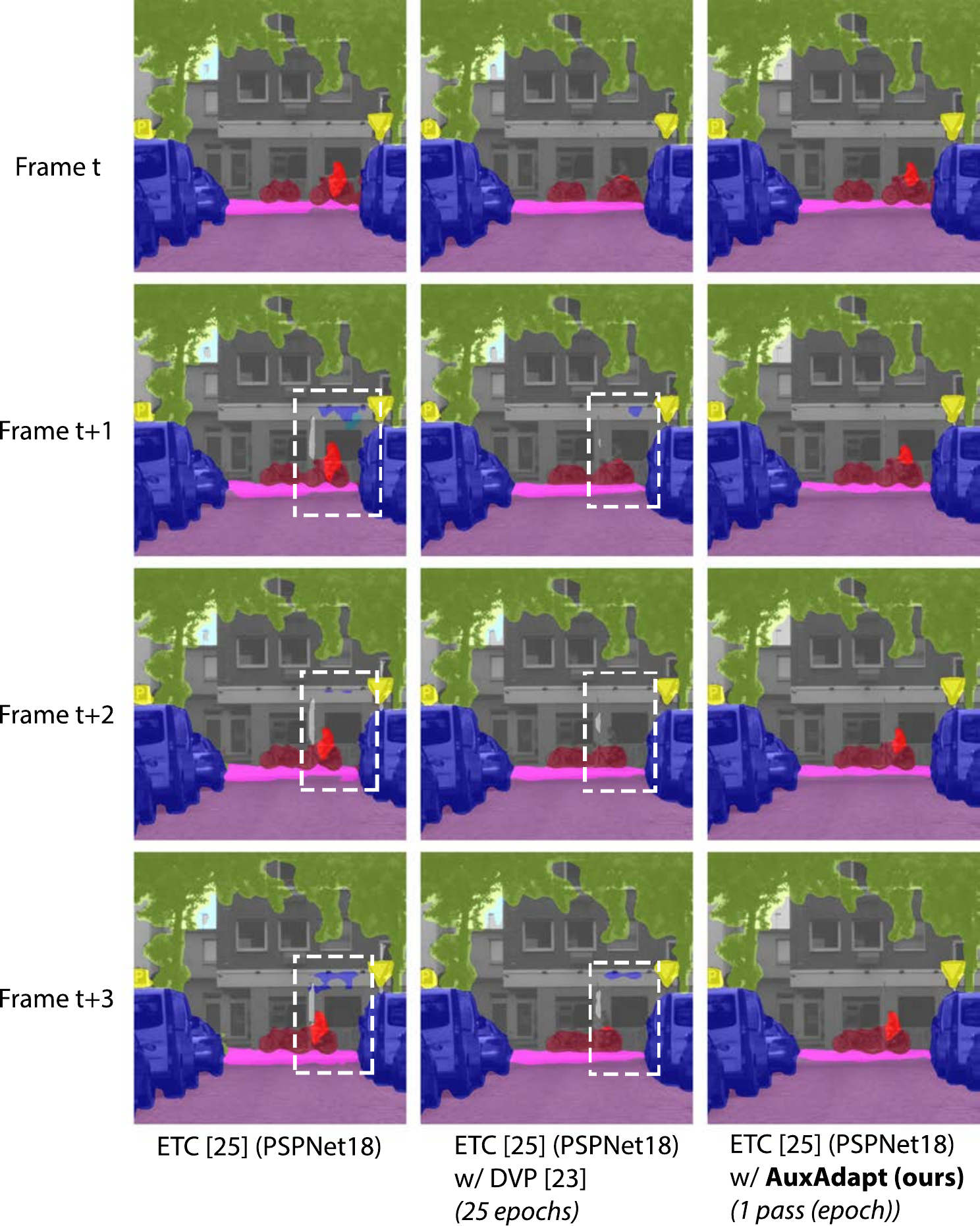}
    \caption{\small Visual results comparing our proposed AuxAdapt with state-of-the-art methods. \textbf{Left:} ETC~\cite{liu2020efficient}. \textbf{Middle:} DVP~\cite{lei2020blind}. \textbf{Right:} AuxAdapt (ours). White dashed boxes highlight temporally inconsistent and inaccurate segmentation. It can be seen that AuxAdapt significantly improves segmentation temporal consistency.} 
\label{fig:sample2}
\vspace{0pt}
\end{figure}

\begin{figure}[t!]
    \centering
    \includegraphics[width=0.95\linewidth]{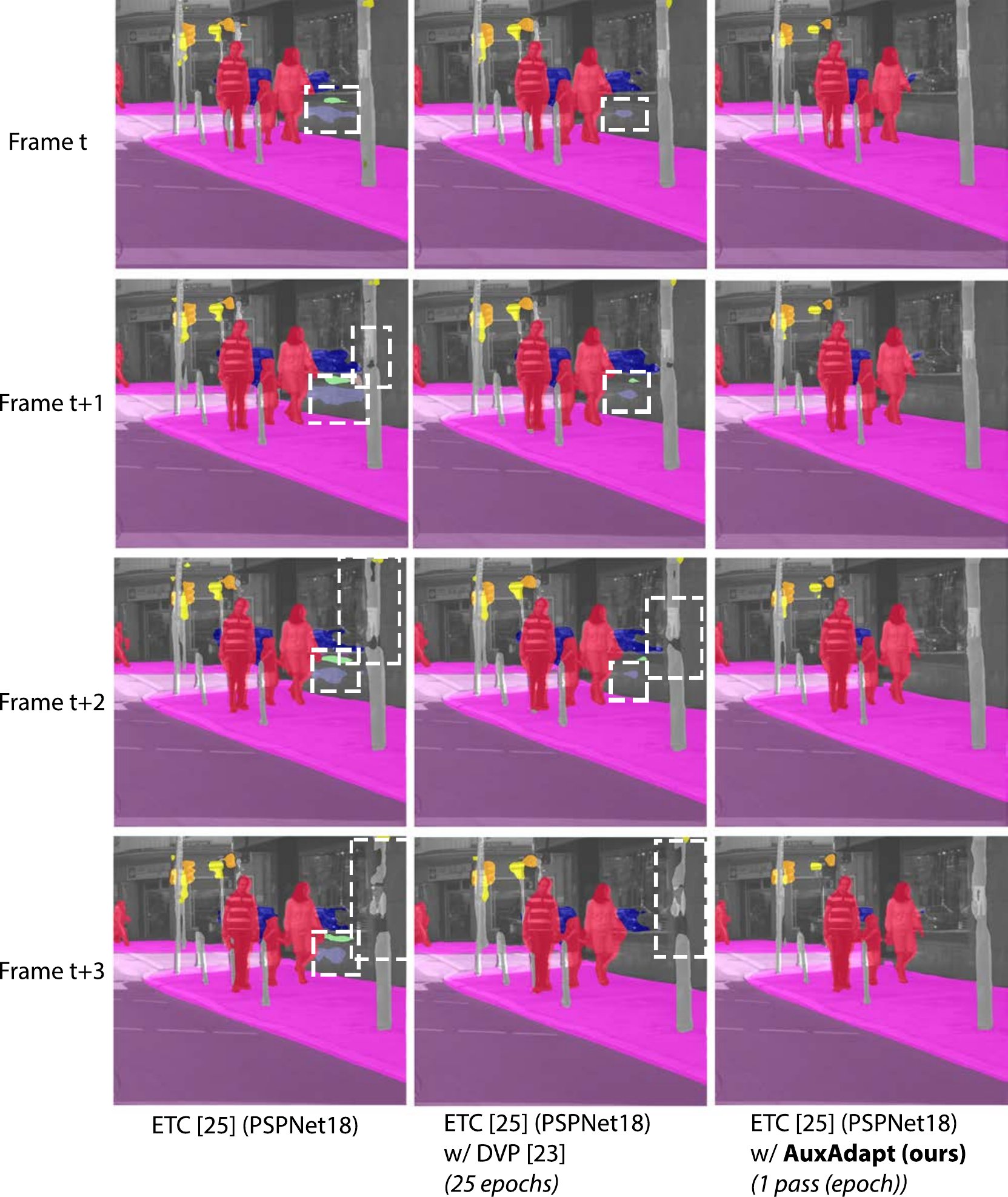}
    \caption{\small Visual results comparing our proposed AuxAdapt with state-of-the-art methods. \textbf{Left:} ETC~\cite{liu2020efficient}. \textbf{Middle:} DVP~\cite{lei2020blind}. \textbf{Right:} AuxAdapt (ours). White dashed boxes highlight temporally inconsistent and inaccurate segmentation. It can be seen that AuxAdapt significantly improves segmentation temporal consistency.} 
\label{fig:sample3}
\vspace{0pt}
\end{figure}

We show visual results to compare AuxAdapt with the state-of-the-art segmentation model, ETC~\cite{liu2020efficient}, which utilizes optical flow during training, and the state-of-the-art test-time adaptation method for improving temporal consistency, DVP~\cite{lei2020blind}. In Fig.~\ref{fig:sample1},~\ref{fig:sample2}, and~\ref{fig:sample3}, it can be seen that AuxAdapt generates temporally more consistent and more accurate video semantic segmentation results. Furthermore, as compared to DVP which requires 25 epochs of training for test-time adaptation, our proposed AuxAdapt requires only one pass of the video.

\end{document}